\documentclass{article}
\usepackage{spconf,amsmath,epsfig}
\usepackage{graphicx}
\usepackage{subcaption}
\usepackage{svg}
\usepackage{soul}

\usepackage{tikz}
\usetikzlibrary{shapes.geometric, arrows}

\title{QUANTIFYING HETEROGENEOUS ECOSYSTEM SERVICES WITH\\ MULTI-LABEL SOFT CLASSIFICATION}

\name{
\begin{tabular}{cc}
Zhihui Tian$^1$, John Upchurch$^2$, G. Austin Simon$^1$, Jos\'e Dubeux$^3$ \\
Alina Zare$^1$, Chang 
 Zhao$^{3*}$, Joel B. Harley$^{1*}$
\end{tabular}
}

\address{\vspace{1.0\baselineskip}$^1$University of Florida, Department of Electrical and Computer Engineering, Gainesville, FL\\
         $^2$University of Florida, Department of Soil, Water, and Ecosystem Sciences, Gainesville, FL \\
         $^3$University of Florida, Department of Agronomy,  Gainesville, FL \\
         $^*$ Chang Zhao and Joel B. Harley are co-senior authors.}

\begin{document}
%
\maketitle
\begin{abstract}
Understanding and quantifying ecosystem services are crucial for sustainable environmental management, conservation efforts, and policy-making. The advancement of remote sensing technology and machine learning techniques has greatly facilitated this process. Yet, ground truth labels, such as biodiversity, are very difficult and expensive to measure. In addition, more easily obtainable proxy labels, such as land use, often fail to capture the complex heterogeneity of the ecosystem. In this paper, we demonstrate how land use proxy labels can be implemented with a soft, multi-label classifier to predict ecosystem services with complex heterogeneity. 
\end{abstract}

%
\section{Introduction}
\label{sec:intro}

Ecosystem services are the benefits nature offers to human societies, such as clean water, crop pollination, or climate regulation \cite{b1}. Ecosystem service quantification enables stakeholders to integrate ecological considerations into economic planning and study the interdependencies between ecosystems and human well-being.


The primary challenge in ecosystem services quantification is that locating and labeling ecosystem services manually is costly, laborious, and limited to a small scale. To address this, ecosystem services quantification often involves two different data processes: (1) transforming remotely sensed information into an easier to obtain proxy variable or label, and (2) turning this proxy variable into ecosystem services.  There are two common types of proxy variables. One type of proxy variable is biophysical variables (such as biomass) that are directly related to an ecosystem service. The second type of proxy variables are indirect variables, such as land use / land cover labels \cite{b2}. Using land use / land cover as a proxy variable involves the prediction of land use based on satellite images. Ecosystem services scores are then assigned to different land use types based on some known relationship \cite{b3}. When using land use as a proxy variable, the accuracy of quantification depends on the spatial resolution of land use / cover classification and the number of classes considered \cite{b4}. 

For land use prediction, two prominent classification methods are pixel-based and object-based \cite{b5}. Pixel-based classification analyzes individual pixels and assigns them to specific land use categories based on spectral information. Object-based classification considers groups of pixels as image objects, which incorporates spatial relationships and contextual information. One advantage of object-based approaches is that changing the classification units from pixels to image objects reduces the variation within classes and eliminates the salt-and-pepper effects commonly seen in pixel-based classification \cite{b6}. Also, a wide range of spatial, textural, and contextual features can be derived as additional information to enhance classification accuracy alongside direct spectral observations. 

Each land use class is linked to specific ecosystem services, taking into account factors such as vegetation cover, soil properties, and hydrological characteristics. One commonly used method to establish this link is through an ecosystem services supply matrix \cite{b3}. This matrix links land use to the ecological integrity (the preservation of ecological systems against general disturbances), ecosystem services supply (the ability of a specific area to provide a particular set of ecosystem goods and services), and ecosystem services demand (the consumption or utilization of ecosystem goods and services within a specific area) \cite{b3}.

\tikzstyle{block} = [rectangle, minimum width=2.5cm, minimum height=1.4cm, text centered, draw=black,text width=2.5cm]
\tikzstyle{arrow} = [thick,->,>=stealth]

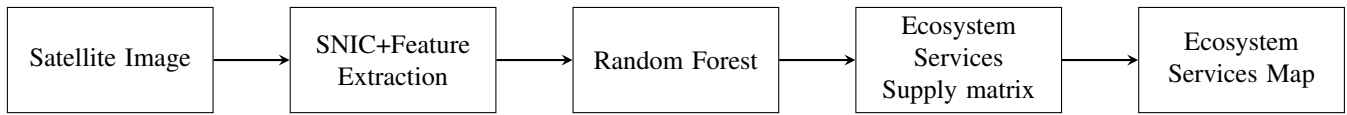
\begin{figure*}[t] 
\centering 
\begin{tikzpicture}[node distance=3.76cm]
\node (step1) [block] {Satellite Image};
\node (step2) [block, right of=step1] {SNIC+Feature Extraction};
\node (step3) [block, right of=step2] {Random Forest};
\node (step4) [block, right of=step3] {Ecosystem Services \\Supply matrix};
\node (step5) [block, right of=step4] {Ecosystem Services Map};
\draw [arrow] (step1) -- (step2);
\draw [arrow] (step2) -- (step3);
\draw [arrow] (step3) -- (step4);
\draw [arrow] (step4) -- (step5);
\end{tikzpicture}
\caption{Processing pipeline for quantifying ecosystem services} 
\label{fig:flowchart} 
\end{figure*}

While land use can be an effective proxy for ecosystem services, it fails to capture the complex heterogeneity of the ecosystem. Pixel-based machine learning algorithms capture heterogeneity by classifying land use based on single-label classification, but these are not suitable for real-world situations since they neglect the non-binary nature of land use and misrepresent spatially local ecosystem relationships, leading to an underrepresentation of ecosystem services provision \cite{b7}. As a result, there is a need for object-based methods that can learn to utilize local ecosystem information while also capturing the ecosystem's heterogeneity. 

In this paper, we present an object-based analysis method that utilizes soft classification to establish heterogeneous ecosystem services scores. Unsupervised learning is first used to transform individual pixels into objects from which we will learn land use proxy labels \cite{b8}. We then achieve a heterogeneity of ecosystem services scores by utilizing soft, multi-label classification \cite{b9}. The predictions from our soft classifier are fed into a fully connected layer encoded with knowledge from an ecosystem services supply matrix. Results demonstrate that our approach allows us to capture  patterns that are hidden when using the proxy labels alone.




\section{METHODOLOGY}
\label{sec:methodology}

We quantify ecosystem services through four steps: 
SNIC (Simple Non-Iterative Clustering), feature extraction, random forest, and a fully connected layer. This pipeline
is shown in Fig.~\ref{fig:flowchart} and described in the following subsections.

\subsection{SIMPLE NON-ITERATIVE CLUSTERING}


The SNIC algorithm is used to segment the satellite images, generating superpixels that we then classify as individual objects \cite{b10}. SNIC segments images by selecting initial cluster centers, refining their centroids based on color and spatial proximity in a single iteration, and producing compact and homogeneous superpixels for image analysis \cite{b11}. 

SNIC offers several advantages over other  segmentation algorithms. It allows for the segmentation of satellite images without the need for labels, which can be difficult to obtain. SNIC is also non-iterative, requiring less memory and is faster compared to other unsupervised segmentation algorithms. More importantly, SNIC groups pixels into homogeneous and coherent regions, facilitating accurate and context-aware classification of image objects. It is important to note that the outputs of the SNIC algorithm do not have the same dimension due to the varying size of superpixels.

\subsection{FEATURE EXTRACTION}

Feature extraction is then applied to each superpixel to give each objects the same dimension. At this step, we extract 48 input features: the minimum, maximum, mean, variance, skewness, and kurtosis of the pixels in each superpixel for each of the eight channels. These features normalize superpixels to the same dimension for classification.

\subsection{LAND USE CLASSIFICATION}

Random forest is an ensemble machine learning algorithm that uses multiple decision trees during training and outputs the most frequently predicted class. Random forest offers key benefits such as its non-parametric nature, high accuracy in classification, and the ability to ascertain the importance of variables \cite{b12}. In this case, random forest is used to train the data processed in the previous steps. Note that the superpixels generated by SNIC can overlap with multiple labels. Under this condition, the random forest is trained with each label separately, using the same input features each time. 

When applying the random forest to test data, we do not select the class with the highest probability. Instead, we consider the random forest as a soft classifier. Each decision tree is trained on a bootstrap sample from the original training dataset and creates a probability distribution over land use classes. These probability distributions are then averaged to generate the final output that we call ensemble probabilities.




\subsection{ECOSYSTEM QUANTIFICATION}

\begin{figure}[t]
\hspace*{1.5em}
  \centering
  \includegraphics[width=0.46\textwidth] {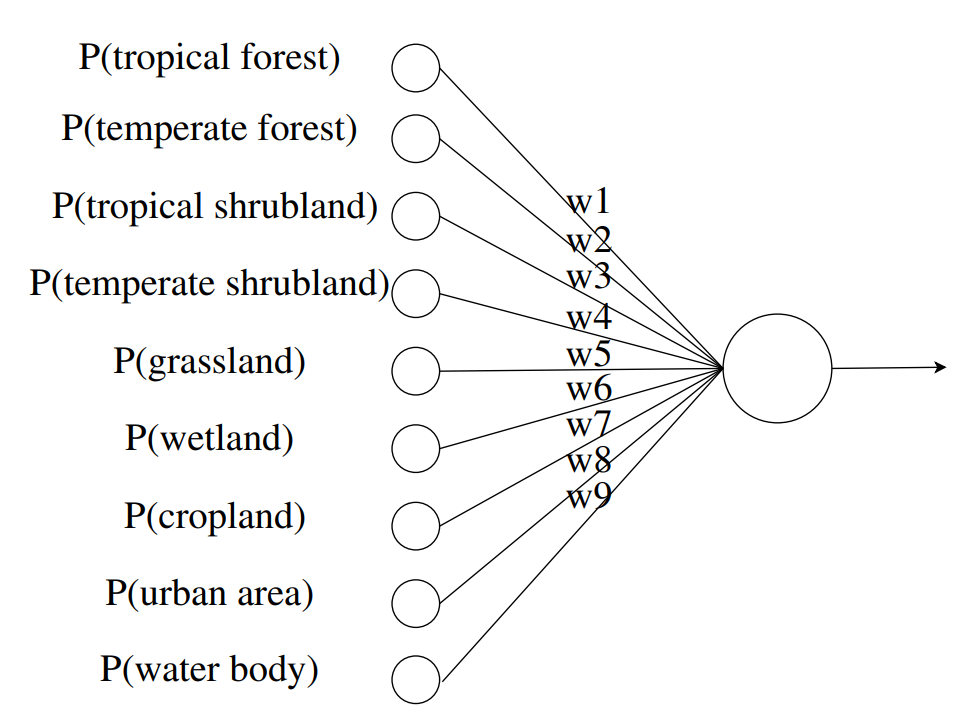}
\caption{The fully connected layer used to translate the land use proxy variable into ecosystem service scores.}
\label{fig:fully_connect}
\end{figure}

In the final step, we utilize a single fully connected layer (Fig.~\ref{fig:fully_connect}), inspired by neural networks, to assess the level of ecosystem services in each superpixel. The input of this fully connected layer is a superpixel's land use probabilities. The weight assigned to each land use class represents their capacities to support an ecosystem service, which is fixed and based on the ecosystem services supply matrix~\cite{b3}. The assessment considers the capacities of land uses to provide specific services according to: 0 = no relevant capacity, 1 = low relevant capacity, 2 = relevant capacity, 3 = medium relevant capacity, 4 = high relevant capacity, and 5 = very high relevant capacity. Each ecosystem service has its own set of weights.  

The output of this layer is the sum of all probabilities multiplied by their respective weights, which provides an ecosystem service capacity score for the respective superpixel. From this, we can create maps of ecosystem service scores (a regression task) rather than of discrete labels (a classification task). This provides a means to compare the differences between areas even if they are the same land use type. For instance, a forest of the same type in different locations will yield varying ecosystem service scores.

\section{DATA SOURCES}

A key assumption of our approach is that a pixel or collection of pixels (i.e., a superpixel) can correspond to some combination of  classes. As a result, we assume that the training labels, which contain only one class per pixel, are approximations of the true multi-class labels. By observing many examples through training and using a soft classifier, we hypothesize that we can predict a more correct multi-label classification.

One consequence of these assumptions is that the training data and testing data can overlap within our dataset. This is because the output of the testing data is not expected to exactly mirror the labels provided to the training data. This is analogous to using the algorithm to ``denoise'' the labels. 

\begin{figure}[t]
\centerline{\includegraphics[width=8cm]{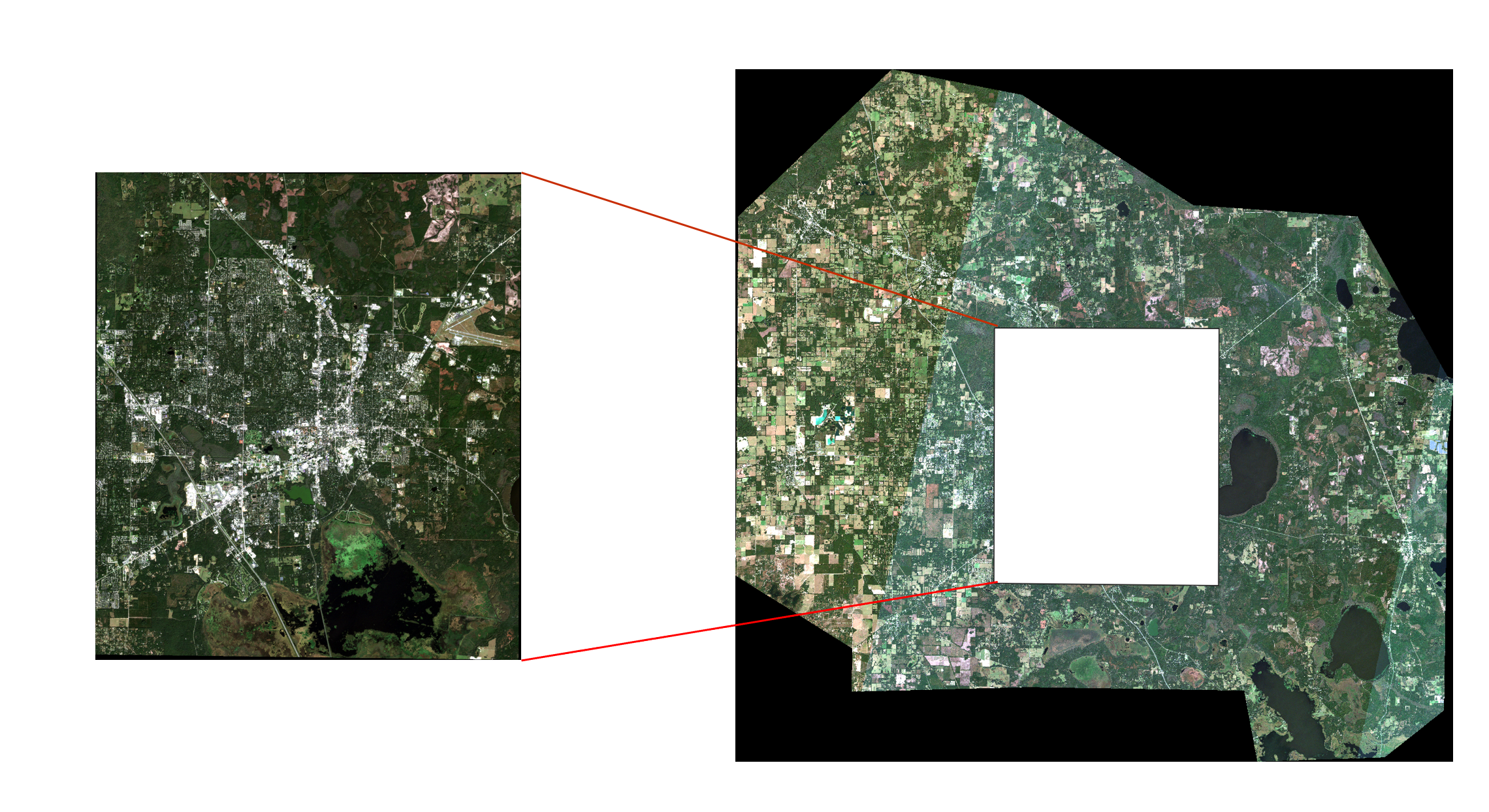}}
\caption{Illustration of the training data (left) and testing data (right) used in predicting ecosystem services scores of Alachua County in Florida, USA.}
\label{fig}
\end{figure}

Our approach is applied to satellite imagery from Planet Labs \cite{b13}, we collect eight-band cloud-free images over Alachua County in Florida, USA. The central portion of the county contains the city of Gainesville and the University of Florida. The entire county image is used as the testing data. The central area of the image is cropped and used as training data, which is shown in Fig.~\ref{fig}. We choose this area for training as it contains a diverse collection of land uses. 

For our land use labels, we use the point sampling function in QGIS software, based on the North American Land Cover, 2020 (Landsat, 30m) \cite{b14}. The limitation of this map is each pixel sampling from the map represents a $30 \times 30$ meter area and is assigned the same label, which is not consistent with the real world, especially for multiple nearby land uses. 


When applying SNIC, we obtain 5000 superpixels for the training and testing data. When applying the random forest, we use an ensemble of 100 estimators. The training labels samples from the land use map is 9000. 


\begin{figure*}[t!b]

  \begin{minipage}{0.3\linewidth}
    \centering
    \includegraphics[width=\linewidth]{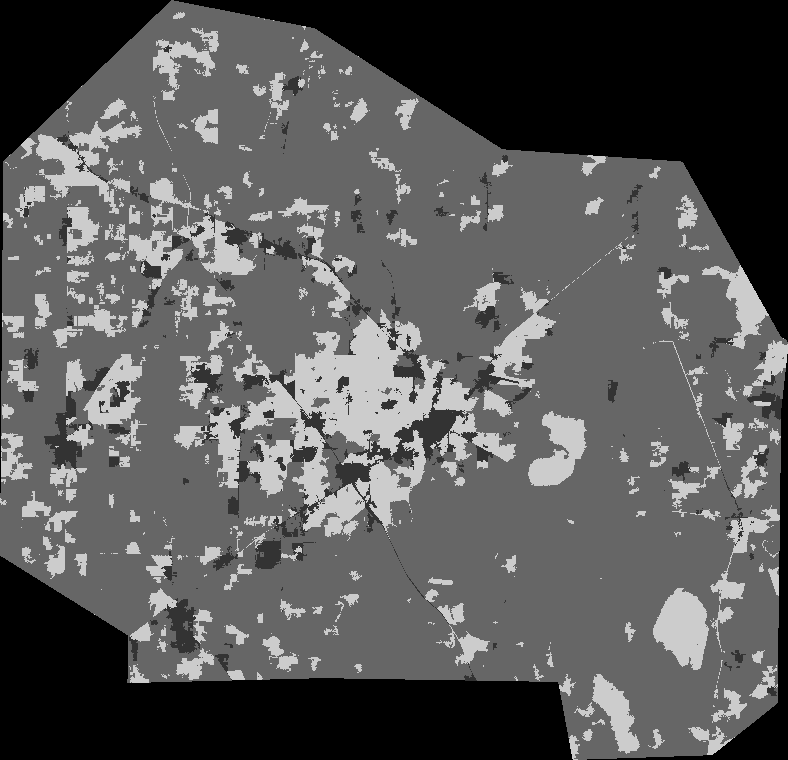}
    \caption*{(\textbf{a}) Biodiversity object-based hard classification}
  \end{minipage}
  \hfill
  \begin{minipage}{0.3\linewidth}
    \centering
    \includegraphics[width=\linewidth]{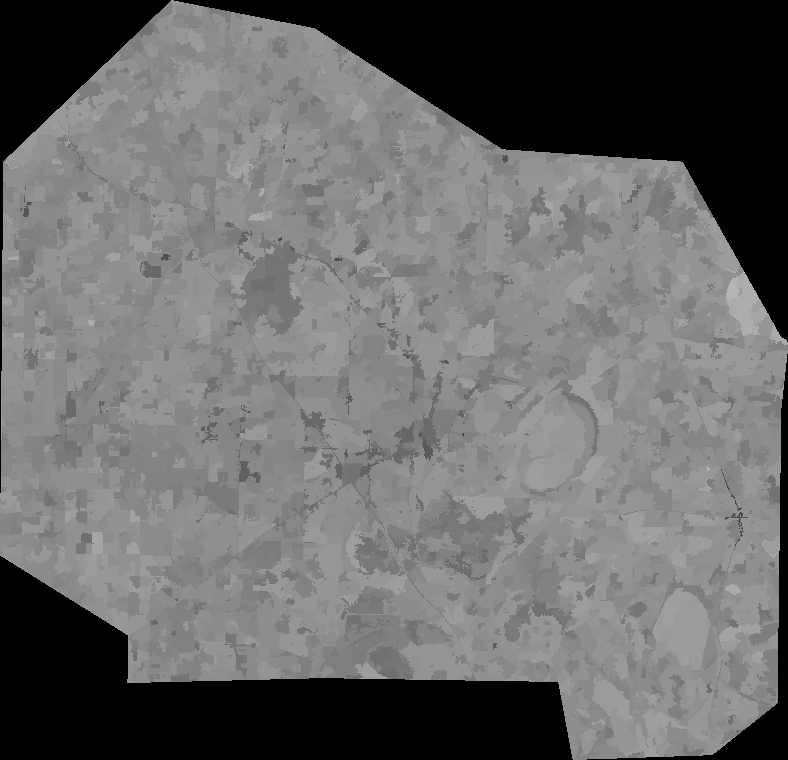}
    \caption*{(\textbf{b}) Biodiversity object-based soft classification (our method)}
  \end{minipage}
  \hfill
  \begin{minipage}{0.3\linewidth}
    \centering
    \includegraphics[width=\linewidth]{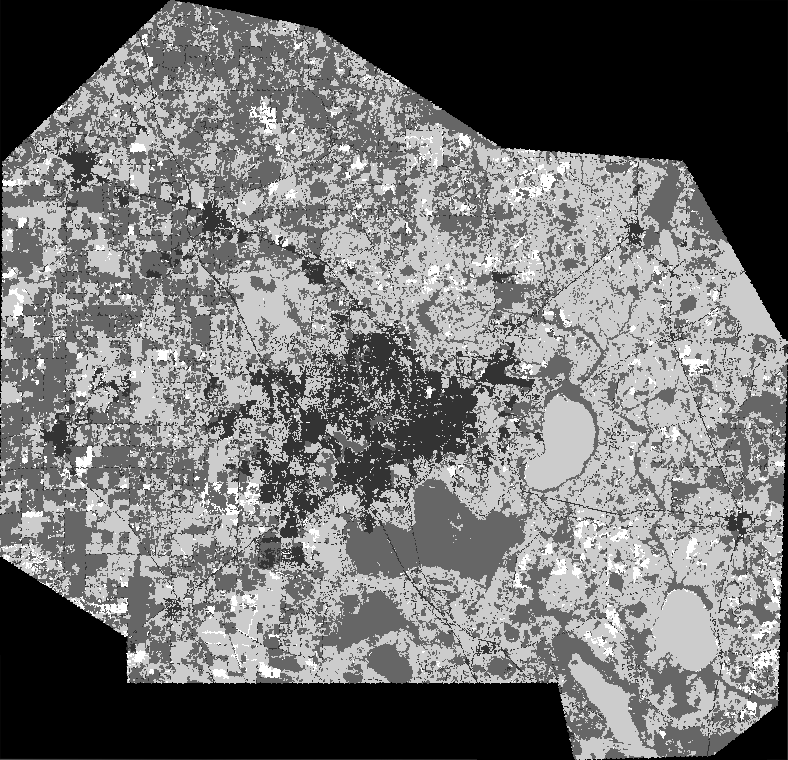}
    \caption*{(\textbf{c}) \hspace{1mm} Biodiversity based on the North American Land Cover Map}
  \end{minipage}

  \vspace{0.3cm} 
  \begin{minipage}{0.3\linewidth}
    \centering
    \includegraphics[width=\linewidth]{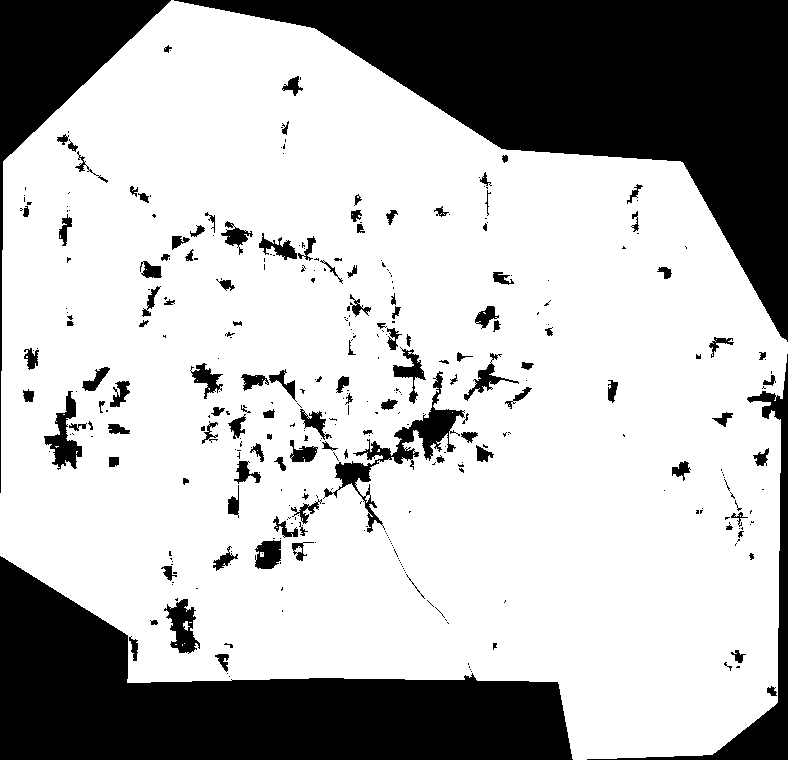}
    \caption*{(\textbf{d}) Groundwater recharge object-based hard classification prediction}
  \end{minipage}
  \hfill
  \begin{minipage}{0.3\linewidth}
    \centering
    \includegraphics[width=\linewidth]{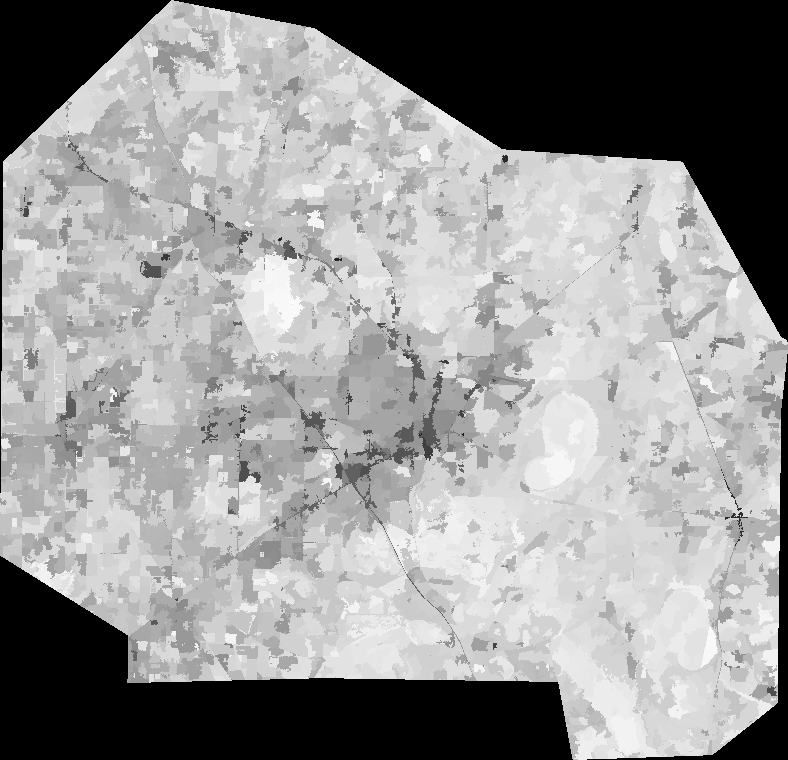}
    \caption*{(\textbf{e}) Groundwater recharge object-based soft classification (our method)}
  \end{minipage}
  \hfill
  \begin{minipage}{0.3\linewidth}
    \centering
    \includegraphics[width=\linewidth]{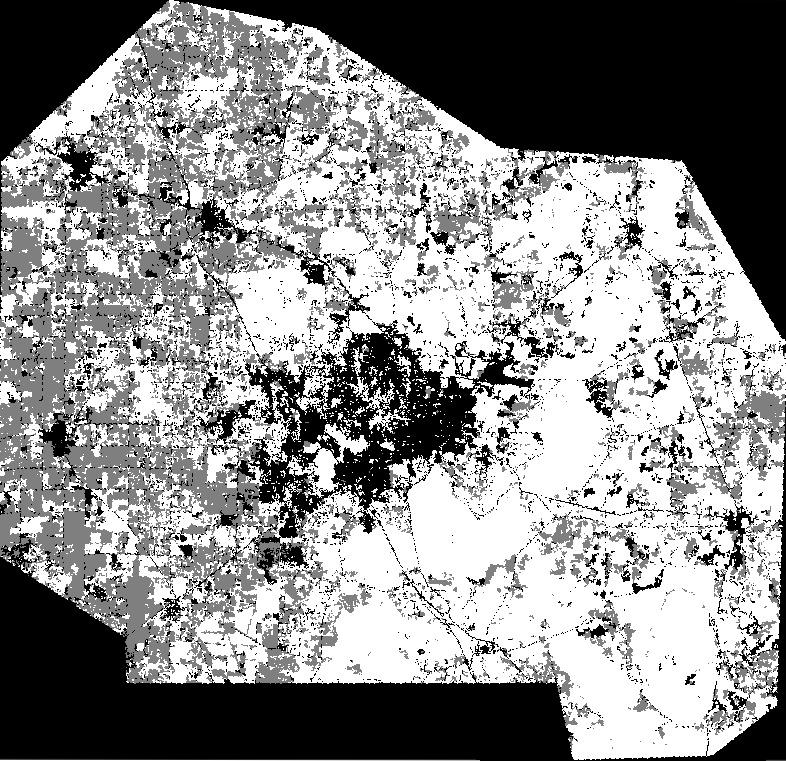}
    \caption*{(\textbf{f}) \hspace{1mm} Groundwater recharge based on the North American Land Cover Map}
  \end{minipage}

  \caption{Illustration of the prediction of two ecosystem services scores, using three different methods. Black represents to lowest score,  while white represents to highest score. The score range for biodiversity is 0-5 and 0-2 for groundwater recharge.}
  \label{fig:results}
\end{figure*}

\section{RESULTS AND DISCUSSION}
\label{sec:pagestyle}

Our method is applied to quantify two distinct ecosystem services: biodiversity and groundwater recharge. The maps resulting from these applications are shown in Fig.~\ref{fig:results}. In these images, the darker areas represent lower ecosystem services scores, whereas the brighter areas indicate higher ecosystem services scores. In addition to our soft classifier (the center column in  Fig.~\ref{fig:results}), we compare our results with that of a hard classifier (i.e., we show ecosystem service scores for the land use determined by the random forest classifier's highest probability class). We also show a pixel-based estimate of the ecosystem services scores corresponding directly to the land use map used for training the random forest classifier. 

Visually, we observe that the three approaches provide very different ecosystem service predictions. The hard classifier predicts the majority of the county as one class. The prediction based on land use map is more diverse spatial but remains limited how it expresses heterogeneity. 

For example, from the soft classifier's groundwater recharge map, we can distinguish densely urban areas (very dark) at the university and downtown Gainesville, the suburban areas (gray) around the city center, and rural/natural areas across the remainder of the county. From the hard classifier, the suburban and natural areas have the same ecosystem service scores. From the land use based map, the urban and suburban areas have the same ecosystem service scores. Similarly, in the biodiversity map, the hard classifier and land use based map provide nearly opposite scores in the suburban areas while the soft classifier provides a more complex prediction that is not at either extreme.

\begin{figure}[h!]
\centering
\includegraphics[width=\columnwidth, clip, trim=0 0cm 1cm 1.2cm]{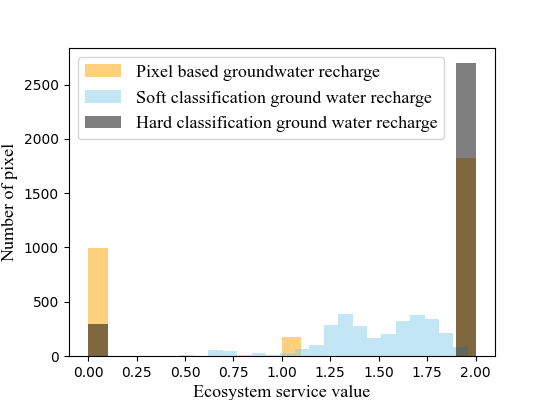}
\caption{{Groundwater recharge score histogram for the three classifiers considered}}
\label{fig:histogram}
\end{figure}

To demonstrate the model’s ability to capture heterogeneity, we analyzed the groundwater recharge scores. Specifically, we extracted the groundwater recharge scores of each map, using the 3000 points randomly sampled points from the training data. We then visualize this data with a histogram. As shown in Fig.~\ref{fig:histogram}, the land use, pixel-based method, and hard classification method both concentrate their values on specific scores (0 or 2, the lowest and highest ecosystem scores for groundwater recharge). The type of distribution is unrealistic as the ecosystem services are expected to vary across locations. In contrast, our soft classification results are more smoothly distributed within the range 0 to 2, which can be interpreted as the changes in geographical location leading to the changes in ecosystem service supply. This demonstrates the model's ability to capture heterogeneity.

\section{CONCLUSIONS}
\label{sec:typestyle}
We have introduced a machine learning model that predicts ecosystem services scores based on satellite imagery, land use proxy variables, and a soft classifier. Object-based classification, which includes an unsupervised algorithm and a random forest, is used to predict a soft land use map. An ecosystem service supply matrix is then used to generate an ecosystem services map, producing continuous ecosystem scores rather than discrete labels. We demonstrated that these maps provide more realistic variations than more traditional strategies.

While this approach shows significant benefits, the results are not yet rigorously validated. Future research will focus on a more rigorous, experimental validation to demonstrate the accuracy of our predictions. 

\end{document}